\documentclass[conference,letterpaper]{IEEEtran}
\IEEEoverridecommandlockouts
\usepackage{graphicx}
\usepackage{bm}
\usepackage{multirow}
\usepackage{amssymb}
\usepackage{amstext}
\usepackage{amsmath}
\usepackage{amsthm}
\usepackage{multicol}
\usepackage{soul}

\usepackage{flushend}

\usepackage{todonotes}
\usepackage{regexpatch}
\makeatletter
\xpatchcmd{\@todo}{\setkeys{todonotes}{#1}}{\setkeys{todonotes}{inline,#1}}{}{}
\makeatother

\usepackage{subfigure}

\newcommand{\vx}{\ensuremath{\boldsymbol{x}}}

\begin{document}

\title{Learning Improved Representations by Transferring Incomplete Evidence Across Heterogeneous Tasks}
\author{
    \IEEEauthorblockN{Athanasios Davvetas and Iraklis A. Klampanos}
\IEEEauthorblockA{\textit{Institute of Informatics and Telecommunications} \\
    \textit{National Centre for Scientific Research ``Demokritos''}\\
    Agia Paraskevi, Athens, Greece \\
\{tdavvetas, iaklampanos\}@iit.demokritos.gr}
}

\maketitle

\begin{abstract}
Acquiring ground truth labels for unlabelled data can be a costly procedure, since it often requires manual labour that is error-prone. Consequently, the available amount of labelled data is increasingly reduced due to the limitations of manual data labelling. It is possible to increase the amount of labelled data samples by performing automated labelling or crowd-sourcing the annotation procedure. However, they often introduce noise or uncertainty in the labelset, that leads to decreased performance of supervised deep learning methods. On the other hand, weak supervision methods remain robust during noisy labelsets \cite{Frenay2014} or can be effective even with low amounts of labelled data \cite{Chapelle2010}. In this paper we evaluate the effectiveness of a representation learning method that uses external categorical evidence called ``Evidence Transfer'', against low amount of corresponding evidence termed as incomplete evidence. Evidence transfer is a robust solution against external unknown categorical evidence that can introduce noise or uncertainty. In our experimental evaluation, evidence transfer proves to be effective and robust against different levels of incompleteness, for two types of incomplete evidence.
\end{abstract}


\section{Introduction}
The availability of large-scale labelled datasets has been attributed as one of the reasons for the increased effectiveness of deep learning. In vision applications, a logarithmic relation between the performance and the volume of training data has been observed \cite{Sun2017}. However, procedures regarding data collection such as acquisition, augmentation and labelling are considered to be a major bottleneck \cite{Roh2018}, since they often require human intervention. In some domains such as semantic detection \cite{Socher2013}, progress is held back by the lack of labelled resources, while other domains have been adapting to learning from limited resources of labelled data \cite{Tajbakhsh2019}.

Augmenting the sample size of labelled data is a proposed way of dealing with insufficient labelled data. Labelsets can be either augmented in automated ways \cite{Tang2018} or by crowd sourcing the labelling procedure \cite{Servadei2018}. Labelset augmentation is often prone to introducing noise. Depending on the preferred augmentation technique, noise could lead to issues such as introducing uncertainty in the decision boundary \cite{Reamaroon2019} or introducing selective bias from disproportionate class sizes.

An alternative way to dealing with insufficient labelled data is using weak labels \cite{Fuentes-Hurtado2019}. Weak labels are usually less informative than ground truth labels. Despite weak labels lacking in information, the procedure of acquiring weak labels for unlabelled data samples is usually less complicated than acquiring ground truth labels. Nevertheless, assuming that weak labels has been acquired for every data sample is not often realistic depending on the nature of the application (e.g real-time prediction), leading to incomplete weak labelsets.

Evidence transfer is a representation learning method that exploits external categorical evidence to manipulate the initial representations of an autoencoder \cite{Davvetas2019}. Learning representations according to external categorical evidence, can be considered as weakly supervised representation learning. External evidence are categorical variables that represent weak labels and not ground truth labels. Additionally, the relation between the dataset and any evidence source is unknown and may introduce uncertainty.

Evidence transfer is effective when introduced with meaningful evidence sources and robust against low quality of evidence. The experimental evaluation of evidence transfer includes individually introducing evidence sources, as well as, combining external categorical evidence sources. Furthermore, it was evaluated against two types of low quality evidence: high uncertainty evidence (random values or white noise evidence) and non-corresponding real evidence (high certainty evidence randomly distributed among samples).

In this paper we investigate whether evidence transfer can be effective and robust against incomplete evidence. We consider two cases of incomplete evidence: uniformly missing samples and selectively missing samples. Selectively missing samples, depending on the level of incompleteness can be considered as low quality evidence, since they can introduce selective bias.

Our contributions are to:
\begin{enumerate}
    \item Evaluate the effectiveness of evidence transfer in a weak supervision setting of missing corresponding samples, known as incomplete supervision
    \item Evaluate the robustness of evidence transfer against two cases of incomplete evidence
    \item Establish evidence transfer as a versatile weak supervision method that can used to exploit both unknown and incomplete external categorical evidence
\end{enumerate}

The rest of the paper is organised as follows. In Section \ref{sec:relwork} we introduce and discuss literature regarding weak supervision and its individual types. We briefly present the background of evidence transfer method and introduce an effective and robust way of using incomplete evidence in Section \ref{sec:evitransf}. We report and discuss the results of the experimental evaluation of incomplete evidence transfer in Section \ref{sec:eval}. In Section \ref{sec:conc} we conclude our work and propose future directions of evidence transfer.

\section{Related Work}
\label{sec:relwork}
Using limited amount of supervised information, that is otherwise known as weak supervision, is a difficult task to define since ``limited'' amount can be interpreted in multiple ways depending on the use case. Limited can be interpreted as containing noise, being incomplete or even be non corresponding. Most frequent types of weak supervision are ``Incomplete Supervision'', ``Inexact Supervision'' and ``Inaccurate Supervision'' \cite{Zhou2017}. Incomplete supervision includes cases during which the amount of labelled data is disproportionate to the amount of available unlabelled data. Inexact supervision includes cases where label information is course-grained, while inaccurate supervision includes cases where labels include errors.

We investigate the ability of evidence transfer to be effective and robust in the incomplete supervision setting, similar to previous inaccurate supervision setting evaluations. Since, evidence transfer does not operate on course-grained evidence, evaluating the performance of evidence transfer in an inexact supervision setting is considered impractical.

The task of semi-supervised learning is a well known case of incomplete supervision. During this task, only a small subset of data samples has available ground truth labels. The dataset is divided into unlabelled and labelled data samples. The objective of most methods is to exploit both unlabelled and labelled data samples. Using a hybrid of generative and discriminative models has been a well-received method in semi-supervised tasks \cite{Zhu2005, Lasserre2006}. Evidence transfer can also be considered as such a method since it uses the layers of an autoencoder both for its generative properties as well as to discriminate the different samples according to evidence. Multiple versions of autoencoders has been used for semi-supervised tasks such as Adversarial Autoencoders \cite{Makhzani2015}, PixelGAN Autoencoders \cite{Makhzani2017} or Variational Autoencoders \cite{Kingma2014}, \cite{Narayanaswamy2017}.

Other generative models such as Generative Adversarial Networks (GAN) \cite{Goodfellow2014} have also been used in hybrid models. From variations of the original GAN called Categorical GAN \cite{Springenberg2015} and Semi-Supervised GAN \cite{Odena2016} to adversarial inference models \cite{Dumoulin2016}, \cite{Belghazi2018}, using genative models as a prior \cite{Jafari2019} and domain specific GANs \cite{Tu2018}. Generative adversarial networks are often evaluated and used in semi-supervised tasks.

Other than creating hybrid methods, self-train methods which are based on meta learning and distinguishing between high noise samples \cite{Sun2019}, \cite{Roli2006} or semi-supervised learning based on graphs or kernels \cite{He2007}, \cite{Liu2010}, \cite{Yang2016}, \cite{Goldberg2006} have also been proposed.

Semi-supervised learning is based on the notion that the available subset of labels corresponds to ground truth labels. Ground truth labels represent the class labels of the dataset. In our case, evidence transfer uses weak labels (external categorical evidence) of unknown relation to the dataset. Incomplete evidence transfer refers to incomplete correspondence between all data samples and weak labels.

Inaccurate supervision includes multiple scenarios of incorrect, non-corresponding or noisy label information. During realistic applications, label noise can be introduced to the labelset either during data acquisition or during automated labelling which can introduce label noise or uncertainty from mislabelled samples \cite{Imoto2019}, \cite{Northcutt2019}, \cite{Brodley2011}. Other inaccurate supervision cases include noisy labels either regarding value noise or semantic noise (e.g fake news) \cite{Wang2014}, \cite{Wang2019}, \cite{Shu2019}, \cite{Helmstetter2018}, \cite{Yao2019}. Lastly, intended biased class proportions \cite{Li2019} or non intended biased class proportions \cite{Wang2013} can also be considered as inaccurate supervision since they introduce selective bias in the labelset.

Despite evidence transfer's previous performance in inaccurate supervision setting, by remaining robust against low quality of evidence, incomplete evidence can be considered as noisy. Selectively missing samples that can occur during acquisition of evidence can introduce selective bias which can impact the outcome of evidence transfer.
 
\section{Evidence Transfer}
\label{sec:evitransf}
\subsection{Background}
Evidence transfer is a two step method. The first step of the method is the initialisation step, followed by the evidence transfer step. During the initialisation, an autoencoder is trained to acquire the baseline latent representations of a primary dataset. The autoencoder is trained as a generative model that learns latent representation which approximate the data generation distribution. During the initial representation learning, no labels are used (ground truth or weak) and therefore is fully unsupervised.

After initialisation, external categorical evidence (weak labels in the form of external auxiliary task, not necessarily derived from or referring to the primary task) are used in order to manipulate the initial learned representations. When introduced with meaningful evidence, evidence transfer, through weak label discrimination, produces manipulated latent representations that are more linearly separable. Increased linear separation is an effect of conditioning initial latent representations to represent the relation between primary data samples and external categorical evidence sources.

In contrast to other methods of representation learning that involve auxiliary variables, evidence transfer avoids the underlying assumption of the constant availability of the auxiliary variables, since in practice, auxiliary variables are either not guaranteed or we may observe the outcome of external processes without having explicit access to the corresponding dataset.

In the context of evidence transfer, any categorical variable or set of categorical variables can be considered as external evidence, as long as it satisfies the assumption that there is a relation between the primary dataset and the categorical variable. The term external or auxiliary refers to the fact that these categorical variables are not necessarily extracted from the primary dataset. They could be outcomes of an auxiliary task that could be performed on the primary dataset or on other unknown auxiliary datasets.

In order to deal with unknown external evidence, evidence transfer was designed to satisfy the following criteria:
\begin{enumerate}
    \item Effectiveness: Evidence transfer should discover and utilise meaningful evidence to effectively manipulate the initial latent representations
    \item Robustness: In case of low quality evidence, evidence transfer step should maintain the initial latent representation quality
    \item Modularity: Evidence transfer should be deployed as an incremental step since evidence availability is not guaranteed
\end{enumerate}

\subsection{Incomplete Evidence Transfer}
Let $\bm{X} = \{\vx^{(1)}, \vx^{(2)}, \dots, \vx^{(N)} \}$ be the primary dataset for representation learning and $\bm{V}=\{\bm{v}^{(1)}, \bm{v}^{(2)}, \dots, \bm{v}^{(M)}\}$ the external categorical evidence. $\bm{V}$ can either be a single set of auxiliary task outcomes or it may contain additional sources noted as $\bm{\mathsf{V}} = \{\bm{V}_1, \bm{V}_2, \dots, \bm{V}_K \}$. In the case of complete evidence, there is full correspondence between each data sample in $\bm{X}$ and in each categorical evidence item in $\bm{V}_i$ with $\bm{V}_i \in \bm{\mathsf{V}}$. In other words, $M=N$ for each $\bm{V}_i$. However, during cases of incomplete evidence $M<N$ across evidence sources. The objective of incomplete evidence transfer is to learn latent representations of $\bm{X}$ according to incomplete evidence $\bm{\mathsf{V}}$ which approximate the effectiveness of complete evidence transfer.

For consistency with incomplete supervision setting notation, let $\bm{X}_{u} = \{\vx^{(1)}_{u}, \vx^{(2)}_{u}, \dots, \vx^{(N-M)}_{u} \} \subset \bm{X}$ be the data samples with no corresponding evidence and $\bm{X}_{l} = \{\vx^{(1)}_{l}, \vx^{(2)}_{l}, \dots, \vx^{(M)}_{l} \} \subset \bm{X}$ with corresponding $\bm{V}=\{\bm{v}^{(1)}, \bm{v}^{(2)}, \dots, \bm{v}^{(M)}\}$ or additional $\bm{\mathsf{V}} = \{\bm{V}_1, \bm{V}_2, \dots, \bm{V}_K \}$.

Denoising autoencoders are used for both phases of evidence transfer. During the initialisation step we train the autoencoder to reconstruct the data samples for all $\bm{X}$, after being corrupted. We use mean squared error between the reconstruction and the primary data samples as defined in Equation \ref{eq:aerecon}.
\begin{equation}\label{eq:aerecon}
    \ell_{AE} = \mathcal{L}(\bm{X}, \bm{X'}) = \frac{1}{N} \sum_{i=1}^{N} (\bm{x}^{(i)}-\bm{x'}^{(i)})^2
\end{equation}
Since the relation between any external evidence source and the primary dataset is unknown, we refrain from using external evidence in its raw form. Instead, we extract latent features from an evidence autoencoder to ensure the robustness of the method, as an intermediate step between initialisation and evidence transfer. Low quality of evidence can be divided into categorical variables with noisy values e.g random values, white noise, uniformly distributed values and categorical variables that introduce decision boundary uncertainty such as non corresponding labels e.g one-hot categorical variable samples introduced in non-corresponding order. White noise evidence is easier to identify by observing the distribution properties of the evidence items. 

In order to create a generic method against any type of evidence (including low quality) we train an uninitialised biased autoencoder to reconstruct each evidence source in $\bm{\mathsf{V}}$. We bias the autoencoder by restricting its generalisation properties through training for a small amount of iterations. Meaningful evidence is characterised by consistency and therefore its distribution can be learned during a small amount of iterations. However, low quality of evidence is characterised by uncertainty or inconsistency that leads to a uniformly distributed latent representations. The objective of the biased evidence autoencoders is defined in Equation \ref{eq:eviaerecon}.

\begin{equation}\label{eq:eviaerecon}
    \ell_{EviAE} = \mathcal{L}(\bm{V}_i, \bm{V}'_i) = \frac{1}{M} \sum_{j=1}^{M} (\bm{v}_i^{(j)}-\bm{v}_i'^{(j)})^2,\ for\ all\ \bm{V}_i \in \bm{\mathsf{V}}
\end{equation}

The evidence transfer step is then deployed using the latent representations acquired by the biased evidence autoencoder instead of raw values. In order to manipulate the initial latent representations of the primary dataset we use cross entropy metric. The asymmetric computation of cross entropy allows the manipulation of latent space according to external evidence by considering the evidence samples as the ``true'' distribution. The intermediate step of pre-processing the evidence samples in combination with the cross entropy loss ensures the robustness criterion of evidence transfer. Meaningful evidence can successfully manipulate the latent space since its representations produce declining values of cross entropy. At the same time, low quality of evidence with high uncertainty representations produces high values of cross entropy.

\begin{figure}
    \includegraphics[width=0.49\textwidth]{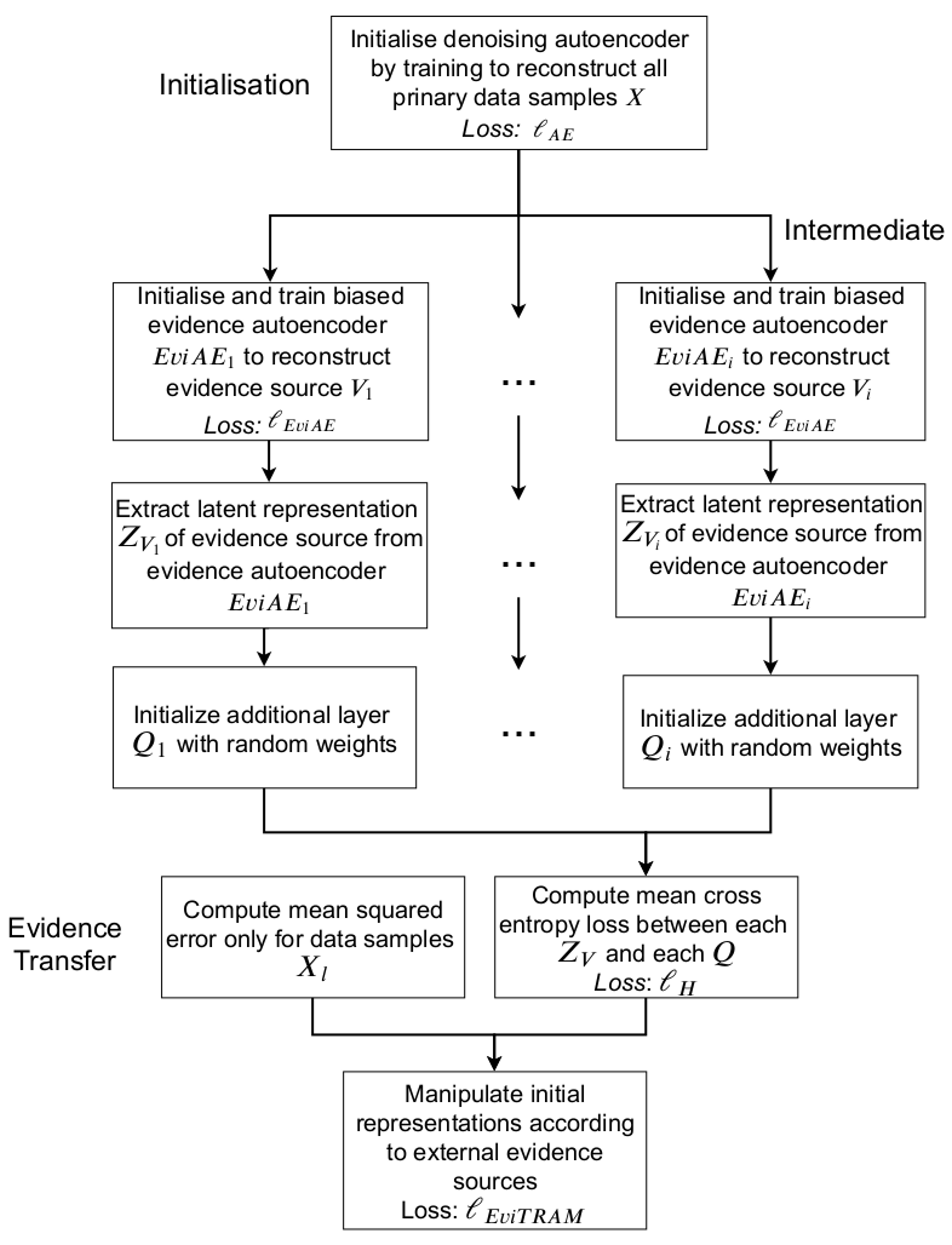}
    \caption{Algorithmic overview of the incomplete evidence transfer algorithm, depicting the various stages of the method. Layers $Q$ are randomly initialised layers that are added in the primary autoencoder after the initialisation. Jointly training uninitialised layers $Q$ along with pretrained layers of the primary autoencoder, using joint objective $\ell_{EviTRAM}$, avoids catastrophic forgetting and finetunes the generalisation performance of the autoencoder even in cases where the amount of $X_{l}$ is low.}
    \label{fig:evitransf_overview}
\end{figure}

\begin{figure}
    \includegraphics[width=0.49\textwidth]{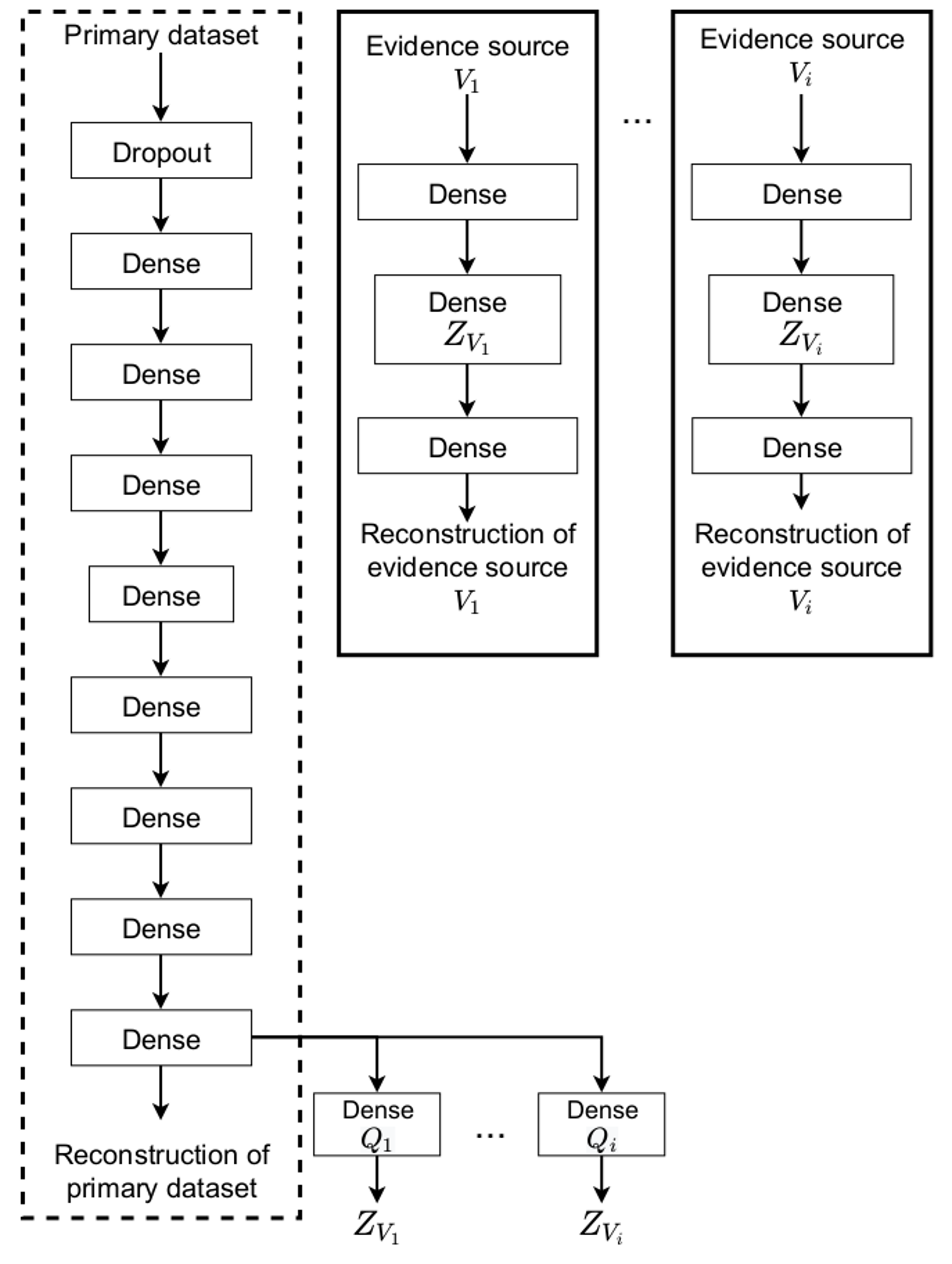}
    \caption{Artificial neural network configuration of incomplete evidence transfer. We depict the initial stacked denoising autoencoder using a dashed line rectangle. Individual evidence autoencoders are depicted using straight line rectangles. Dense boxes represent fully connected layers, we use ReLU activations for all layers except the bottleneck of the autoencoder. The bottleneck has a linear activation. We use Softmax activation both for the layers of the evidence autoencoder, as well as, for additional layers $Q_i$.}
    \label{fig:model_overview}
\end{figure}

To reject evidence of low quality, we incorporate the representations acquired from the evidence autoencoder by using new additional uninitialised layers in our primary autoencoder. The cross entropy is computed between the output of additional layers $\textnormal{Q} = \{\textnormal{Q}_1, \textnormal{Q}_2, \dots, \textnormal{Q}_K \}$ and evidence representations $\textnormal{Z}_{\bm{\mathsf{V}}} = \{\textnormal{Z}_{V_1}, \textnormal{Z}_{V_2}, \dots, \textnormal{Z}_{V_K} \}$, as defined in Equation \ref{eq:crossent}. We cooperatively train the primary autoencoder to manipulate its latent representations by using both the original reconstruction objective and the mean cross entropy of $\bm{\mathsf{V}}$, with all $\bm{V}_i$ being treated equally. The objective of evidence transfer step is defined in Equation \ref{eq:evitram}. An algorithmic overview of incomplete evidence transfer is depicted in Figure \ref{fig:evitransf_overview}, while Figure \ref{fig:model_overview} depicts the artificial neural network configuration of evidence transfer. During low quality evidence, cross entropy loss produces high values that gradually decay the weights of layers $\textnormal{Q}$, allowing reconstruction error to return the latent space to its initial version. 

Introducing new uninitialised layers $\textnormal{Q}$ additionally benefits evidence transfer by avoiding catastrophic forgetting. Evidence transfer belongs in the ``joint optimization'' methods that do not suffer from catastrophic forgetting \cite{Li2018}. This means that after training with all the data samples, explicitly training evidence transfer objective with only $\bm{X}_{l}$ samples will not restrict the generalisation of learning representations for $\bm{X}$. Contrarily, it is considered as finetuning the pretrained layers to minimise the new objective of evidence transfer.

\begin{equation}\label{eq:crossent}
    \ell_{H} = \frac{1}{K} \sum_{j=1}^{K}H(\textnormal{Z}_{V_j}, \textnormal{Q}_j)
\end{equation}

\begin{equation}\label{eq:evitram}
    \ell_{EviTransf} = \underbrace{\frac{1}{N} \sum_{i=1}^{N} (\bm{x}_{l}^{(i)}-\bm{x'}^{(i)}_{l})^2}_{\text{mean squared error of $\bm{X}_l$}} + \lambda * \ell_{H}
\end{equation}

\begin{table*}[h]
\caption{Experimental evaluation of evidence transfer for the MNIST dataset. $W$ represents the width of the categorical variable that is synonymous to the number of classes of the auxiliary task. $N$ represents the samples size of evidence compare to $M$ primary dataset size. For class incomplete evidence, the left side of ``$\rightarrow$'' represents the original number of classes of the auxiliary task, with the right side representing only the available selected number of auxiliary task classes. Real evidence of width 3 corresponds to the $y\ mod\ 3$ relation, while width 4 corresponds to $hash(y)\ mod\ 4$. Width 10 evidence is the labelset of MNIST.}
\label{tab:mnist} \centering %
\subtable[Sample percent incompleteness]{
    \begin{tabular}{|c|c|c|}
    \hline 
    \multirow{1}{*}{Configuration} & ACC (\%) & NMI (\%)\tabularnewline
    \hline 
    Baseline & 82.03 & 76.25\tabularnewline
    \hline 
    Real evidence (w: 3, M=0.1 {*} N) & 82.90 (+0.87) & 76.84 (+0.35)\tabularnewline
    \hline 
    Real evidence (w: 3, M=0.3 {*} N) & 91.23 (+9.20) & 82.93(+6.68)\tabularnewline
    \hline 
    Real evidence (w: 3, M=N) & 95.57 (+13.54) & 89.59 (+13.34)\tabularnewline
    \hline 
    Real evidence (w: 4, M=0.1 {*} N) & 89.83 (+7.80) & 81.14 (+4.89)\tabularnewline
    \hline 
    Real evidence (w: 4, M=0.3 {*} N) & 94.74 (+12.71) & 87.91 (+11.66)\tabularnewline
    \hline 
    Real evidence (w: 4, M=N) & 96.40 (+14.37) & 91.10 (+14.85)\tabularnewline
    \hline 
    Real evidence (w: 10, M=0.1 {*} N) & 84.02 (+1.99) & 78.00 (+1.75)\tabularnewline
    \hline 
    Real evidence (w: 10, M=0.3 {*} N) & 94.57 (+12.54) & 87.68 (+11.43)\tabularnewline
    \hline 
    Real evidence (w: 10, M=N) & 96.71 (+14.68) & 91.77 (+15.52)\tabularnewline
    \hline 
     2 Real evidence (w: 3,4 - M=0.1 {*} N) & 82.99 (+0.96) & 77.02 (+0.77)\tabularnewline
    \hline 
     2 Real evidence (w: 3,4 - M=0.3 {*} N) & 93.11 (+11.08) & 85.39 (+9.14)\tabularnewline
    \hline 
     2 Real evidence (w: 3,4 - M=N) & 97.72 (+15.69) & 93.93 (+17.68)\tabularnewline
    \hline 
    \end{tabular}
}
\subtable[Class incompleteness/biasing]{
    \begin{tabular}{|c|c|c|}
    \hline 
    \multirow{1}{*}{Configuration} & ACC (\%) & NMI (\%)\tabularnewline
    \hline 
    Baseline & 82.03 & 76.25\tabularnewline
    \hline 
    Real evidence (w: 3) & 95.57 (+13.54) & 89.59 (+13.34)\tabularnewline
    \hline 
    Real evidence (w: 3 $\rightarrow$ 2) & 90.32 (+8.29) & 82.19 (+5.94)\tabularnewline
    \hline 
    Real evidence (w: 3 $\rightarrow$ 1) & 82.38 (+0.35) & 76.60(+0.35)\tabularnewline
    \hline 
    Real evidence (w: 4) & 96.40 (+14.37) & 91.10 (+14.85)\tabularnewline
    \hline 
    Real evidence (w: 4 $\rightarrow$ 3) & 92.09 (+10.06) & 86.21 (+9.96)\tabularnewline
    \hline 
    Real evidence (w: 4 $\rightarrow$ 2) & 86.56 (+4.53) & 80.42 (+4.17)\tabularnewline
    \hline 
    Real evidence (w: 10) & 96.71 (+14.68) & 91.77 (+15.52)\tabularnewline
    \hline 
    Real evidence (w: 10 $\rightarrow$ 9) & 96.27 (+14.24) & 91.34 (+15.09)\tabularnewline
    \hline 
    Real evidence (w: 10 $\rightarrow$ 8) & 95.77 (+13.74) & 90.30 (+14.05)\tabularnewline
    \hline 
    2 Real evidence (w: 3,4) & 97.72 (+15.69) & 93.93 (+17.68)\tabularnewline
    \hline 
     2 Real evidence (w: 3,4 $\rightarrow$ 2,3) & 90.22 (+8.19) & 81.96 (+5.71)\tabularnewline
    \hline 
     2 Real evidence (w: 3,4 $\rightarrow$ 1,2) & 82.36 (+0.33) & 76.65 (+0.40)\tabularnewline
    \hline 
    \end{tabular}
}
\end{table*}

\begin{table*}[h]
\caption{Experimental evaluation of evidence transfer for the 20newsgroups dataset. $W$ represents the width of the categorical variable that is synonymous to the number of classes of the auxiliary task. $N$ represents the samples size of evidence compare to $M$ primary dataset size. For class incomplete evidence, the left side of ``$\rightarrow$'' represents the original number of classes of the auxiliary task, with the right side representing only the available selected number of auxiliary task classes. Most of the evidence in 20newgroups represents a superset of the original labels. Real evidence with width 20 corresponds to the labelset of 20newsgroups. Real evidence with width 5 is a superset consisting of the topics: ``Computers'', ``Recreational'', ``Science'', ``Talk'' and ``Misc'' topics. Width 6 evidence consists of ``Sports'', ``Politics'', ``Religion'', ``Vehicles'', ``Systems'' and ``Science''} 
\label{tab:20ng} \centering %
\subtable[Sample percent incompleteness]{
    \begin{tabular}{|c|c|c|}
    \hline 
    \multirow{1}{*}{Configuration} & ACC (\%) & NMI (\%)\tabularnewline
    \hline 
    Baseline & 21.57 & 25.01\tabularnewline
    \hline 
    Real evidence (w: 5, M=0.1 {*} N) & 23.59 (+2.02) & 29.27 (+4.26)\tabularnewline
    \hline 
    Real evidence (w: 5, M=0.3 {*} N) & 30.42 (+8.85) & 39.21 (+14.21)\tabularnewline
    \hline 
    Real evidence (w: 5, M=N) & 34.18 (+12.61) & 57.35 (+32.34)\tabularnewline
    \hline 
    Real evidence (w: 6, M=0.1 {*} N) & 25.03 (+3.46) & 31.04 (+6.03)\tabularnewline
    \hline 
    Real evidence (w: 6, M=0.3 {*} N) & 34.04 (+12.47) & 41.71 (+16.70)\tabularnewline
    \hline 
    Real evidence (w: 6, M=N) & 32.78 (+11.21) & 60.15 (+35.15)\tabularnewline
    \hline 
    Real evidence (w: 20, M=0.1 {*} N) & 24.33 (+2.76) & 27.38 (+2.38)\tabularnewline
    \hline 
    Real evidence (w: 20, M=0.3 {*} N) & 54.92 (+33.35) & 49.94 (+24.94)\tabularnewline
    \hline 
    Real evidence (w: 20, M=N) & 88.90 (+67.33) & 90.01 (+65.00)\tabularnewline
    \hline 
     2 Real evidence (w: 5,6 - M=0.1 {*} N) & 27.04 (+5.47) & 33.22 (+8.21)\tabularnewline
    \hline 
     2 Real evidence (w: 5,6 - M=0.3 {*} N) & 36.48 (+14.92) & 44.75 (+19.75)\tabularnewline
    \hline 
     2 Real evidence (w: 5,6 - M=N) & 46.19 (+24.62) & 68.31 (+43.30)\tabularnewline
    \hline 
    \end{tabular}
}
\subtable[Class incompleteness/biasing]{
    \begin{tabular}{|c|c|c|}
    \hline 
    \multirow{1}{*}{Configuration} & ACC (\%) & NMI (\%)\tabularnewline
    \hline 
    Baseline & 21.57 & 25.01\tabularnewline
    \hline 
    Real evidence (w: 5) & 34.18 (+12.61) & 57.35 (+32.34)\tabularnewline
    \hline 
    Real evidence (w: 5 $\rightarrow$ 4) & 31.27 (+9.70) & 49.01 (+24.00)\tabularnewline
    \hline 
    Real evidence (w: 5 $\rightarrow$ 3) & 25.95 (+4.38) & 35.59 (+10.58)\tabularnewline
    \hline 
    Real evidence (w: 6) & 32.78 (+11.21) & 60.15 (+35.15)\tabularnewline
    \hline 
    Real evidence (w: 6 $\rightarrow$ 5) & 30.43 (+8.87) & 49.53 (+24.52)\tabularnewline
    \hline 
    Real evidence (w: 6 $\rightarrow$ 4) & 25.21 (+3.64) & 36.41 (+11.41)\tabularnewline
    \hline 
    Real evidence (w: 20) & 88.90 (+67.33) & 90.01 (+65.00)\tabularnewline
    \hline 
    Real evidence (w: 20 $\rightarrow$ 19) & 79.55 (+57.99) & 83.40 (+58.39)\tabularnewline
    \hline 
    Real evidence (w: 20 $\rightarrow$ 18) & 76.65 (+55.08) & 80.20 (+55.19)\tabularnewline
    \hline 
    2 Real evidence (w: 5,6) & 46.19 (+24.62) & 68.31 (+43.30)\tabularnewline
    \hline 
     2 Real evidence (w: 5,6 $\rightarrow$ 4,5) & 21.56 (-0.01) & 39.03 (+14.02)\tabularnewline
    \hline 
     2 Real evidence (w: 5,6 $\rightarrow$ 3,4) & 24.44 (+2.88) & 30.81 (+5.81)\tabularnewline
    \hline 
    \end{tabular}
}
\end{table*}

\section{Evaluation and Results}
\label{sec:eval}
\subsection{Experimental setup}
\label{subsec:setup}
\subsubsection{Datasets}
We evaluate incomplete evidence transfer both on image and text datasets. We use MNIST dataset that contains handwritten digits with 10 class labels varying from 0 to 9, as well as, CIFAR-10 dataset \cite{cifar10} that contains RGB images depicting different vehicles or animals (e.g airplane, horse, etc.). For the CIFAR-10 experiments we use features extracted from a pretrained VGG-16 \cite{vgg} network on ImageNet \cite{imagenet} instead of raw images. 

Furthermore, we use 20newsgroups dataset that contains articles that can be classified into 20 news topics, as well as Reuters Corpus Volume I (RCV1) \cite{Lewis2004} that also contains articles that can be classified into 103 categories (4 root categories with additional sub-categories). To achieve consistency with the other three datasets, we created and used a subset of RCV1 with 10 categories (4 root categories plus 6 sub categories) and 96,933 data samples. To train our models for the 20newsgroups dataset, we use extracted features from a pretrained word2vec model \cite{Mikolov} on, the Google News Corpus. During training of Reuters100k subset we used TF-IDF features.
\subsubsection{Evidence}
We simulated two types of incomplete evidence: sample percent incompleteness and class incompleteness. Sample percent incompleteness is simulated by uniformly removing a percent of the complete evidence set. In other words, the evidence classes are all represented in the evidence set with fewer samples. Class incompleteness is simulated by removing an amount of classes from the complete evidence set, i.e remove all samples of one class from the evidence set. Incomplete class evidence can be seen as a case where evidence is still in the process of gathering and therefore some classes are missing. For all experiments we use incomplete yet corresponding real evidence. For all datasets we experiment with one and two sources of evidence. For CIFAR-10 we additionally experimented with three sources of evidence.
\subsubsection{Metrics}
We quantitatively evaluate the robustness and effectiveness of incomplete evidence transfer by measuring its performance on the task of clustering the primary dataset samples. We perform clustering before and after applying various stages of incomplete evidence in order to measure their distance from the baseline solution. Baseline solution represents clustering the latent representations acquired during the initialisation phase, i.e before applying incomplete evidence transfer. The metrics that we are using during our experiments are Unsupervised Clustering Accuracy (ACC) and Normalised Mutual Information (NMI).

\begin{table*}[h]
\caption{Experimental evaluation of evidence transfer for the Reuters-100k dataset. $W$ represents the width of the categorical variable that is synonymous to the number of classes of the auxiliary task. $N$ represents the samples size of evidence compare to $M$ primary dataset size. For class incomplete evidence, the left side of ``$\rightarrow$'' represents the original number of classes of the auxiliary task, with the right side representing only the available selected number of auxiliary task classes. Width 4 evidence is the root categories of RCV1. Width 5 is a re-categorisation of 10 subcategories into 5 groups. Width 10 is the subcategory labelset.}
\label{tab:reu100k} \centering %
\subtable[Sample percent incompleteness]{
    \begin{tabular}{|c|c|c|}
    \hline 
    \multirow{1}{*}{Configuration} & ACC (\%) & NMI (\%)\tabularnewline
    \hline 
    Baseline & 41.13 & 32.72\tabularnewline
    \hline 
    Real evidence (w: 4, M=0.1 {*} N) & 41.66 (+0.53) & 32.98 (+0.26)\tabularnewline
    \hline 
    Real evidence (w: 4, M=0.3 {*} N) & 44.48 (+3.35) & 36.12 (+3.41)\tabularnewline
    \hline 
    Real evidence (w: 4, M=N) & 43.34 (+2.21) & 36.24 (+3.52)\tabularnewline
    \hline 
    Real evidence (w: 5, M=0.1 {*} N) & 44.51 (+3.39) & 35.84 (+3.13)\tabularnewline
    \hline 
    Real evidence (w: 5, M=0.3 {*} N) & 42.98 (+1.86) & 33.28 (+0.56)\tabularnewline
    \hline 
    Real evidence (w: 5, M=N) & 47.00 (+5.87) & 38.75 (+6.03)\tabularnewline
    \hline 
    Real evidence (w: 10, M=0.1 {*} N) & 45.47 (+4.35) & 37.18 (+4.46)\tabularnewline
    \hline 
    Real evidence (w: 10, M=0.3 {*} N) & 46.18 (+5.05) & 36.86 (+4.14)\tabularnewline
    \hline 
    Real evidence (w: 10, M=N) & 48.27 (+7.14) & 41.23 (+8.51)\tabularnewline
    \hline 
     2 Real evidence (w: 4,5 - M=0.1 {*} N) & 45.30 (+4.17) & 36.91 (+4.19)\tabularnewline
    \hline 
     2 Real evidence (w: 4,5 - M=0.3 {*} N) & 48.57 (+7.45) & 38.01 (+5.29)\tabularnewline
    \hline 
     2 Real evidence (w: 4,5 - M=N) & 50.54 (+9.41) & 41.81 (+9.09)\tabularnewline
    \hline 
    \end{tabular}
}
\subtable[Class incompleteness/biasing]{
    \begin{tabular}{|c|c|c|}
    \hline 
    \multirow{1}{*}{Configuration} & ACC (\%) & NMI (\%)\tabularnewline
    \hline 
    Baseline & 41.13 & 32.72\tabularnewline
    \hline 
    Real evidence (w: 4) & 43.34 (+2.21) & 36.24 (+3.52)\tabularnewline
    \hline 
    Real evidence (w: 4 $\rightarrow$ 3) & 41.63 (+0.50) & 32.75 (+0.03)\tabularnewline
    \hline 
    Real evidence (w: 4 $\rightarrow$ 2) & 41.31 (+0.18) & 32.17 (-0.54)\tabularnewline
    \hline 
    Real evidence (w: 5) & 47.00 (+5.87) & 38.75 (+6.03)\tabularnewline
    \hline 
    Real evidence (w: 5 $\rightarrow$ 4) & 46.53 (+5.40) & 41.86 (+9.14)\tabularnewline
    \hline 
    Real evidence (w: 5 $\rightarrow$ 3) & 41.99 (+0.87) & 40.87 (+8.15)\tabularnewline
    \hline 
    Real evidence (w: 10) & 48.27 (+7.14) & 41.23 (+8.51)\tabularnewline
    \hline 
    Real evidence (w: 10 $\rightarrow$ 9) & 59.41 (+18.28) & 49.83 (+17.12)\tabularnewline
    \hline 
    Real evidence (w: 10 $\rightarrow$ 8) & 58.39 (+17.27) & 49.84 (+17.12)\tabularnewline
    \hline 
    2 Real evidence (w: 4,5) & 50.54 (+9.41) & 41.81 (+9.09)\tabularnewline
    \hline 
    2 Real evidence (w: 4,5 $\rightarrow$ 3,4) & 41.32 (+0.19) & 32.54 (-0.18)\tabularnewline
    \hline 
     2 Real evidence (w: 4,5 $\rightarrow$ 2,3) & 41.46 (+0.33) & 32.82 (+0.10)\tabularnewline
    \hline 
    \end{tabular}
}
\end{table*}

\subsection{Discussion of results}
As observed in Tables \ref{tab:mnist}, \ref{tab:20ng}, \ref{tab:reu100k}, \ref{tab:cifar} introducing incomplete evidence in evidence transfer does not affect the performance metrics of the robustness or effectiveness criteria. Our experiments show that percent incomplete evidence is mostly effective. The effectiveness gain of percent incomplete evidence is equivalent to the amount of missing samples. Therefore, using incomplete evidence that represents all classes, always leads to performance gain which is equivalent to the number of samples in each class. 

On the other hand, class incomplete evidence (limited class proportions) in some experiments approaches the performance of low quality evidence. During experiments where the evidence classes are low (e.g evidence with 3 classes), having no samples representing one or two classes is heavily biasing the generalisation performance of evidence transfer. In cases where the amount of evidence classes is high (e.g evidence with 10 classes), limited class proportions are not as heavily biasing and are equivalently effective.

Despite some cases of class incomplete evidence behaving as low quality, there is no significant decrease in effectiveness from the baseline solution. From our experiments we can conclude that using unknown evidence, that may be incomplete either in the amount of representative samples of each class or in class proportions does not decrease the initial performance and may lead to significant gain in effectiveness.

Experimental evaluations in MNIST (Table \ref{tab:mnist}) confirm the general notion of uniformly incomplete evidence being proportionally effective. The performance of selectively incomplete evidence, relies on the total amount of auxiliary task classes. The increase in performance becomes more significant with the increase of total auxiliary task classes. Experimental evaluations in 20newsgroups (Table \ref{tab:20ng}) and CIFAR-10 (Table \ref{tab:cifar}) is consistent in the same way as MNIST. 

Evaluation with Reuters-100k (Table \ref{tab:reu100k}), while fairly consistent with the other evaluations, during selectively incomplete evidence explicit optimisation was required to achieve robustness due to the intrinsic properties of the labelset. As mentioned during Subsection \ref{subsec:setup}, the structure of RCV1 labels are derived from 4 root categories. The same structure is preserved in our subset Reuters-100k using 4 root categories along with 6 sub-categories. This intrinsic structure is prone to selective bias during selectively incomplete evidence, therefore requiring explicit optimisation of the evidence transfer objective.

\section{Conclusions and Future Work}
\label{sec:conc}
In this paper we evaluated the effectiveness and robustness of evidence transfer in the weak supervision setting of incomplete supervision. Evidence transfer proved to be both effective and robust during experimental evaluation with two types of incomplete evidence, as well as, introducing multiple sources of incomplete evidence. Incomplete evidence was simulated both by uniformly and selectively reducing the class proportion samples. From the conducted experiments we can conclude that evidence transfer works as an all around weak supervision method of learning representations with lacking primary dataset labels.

Although during experimental evaluation we tried to simulate realistic cases of weak supervision, there is a need of evaluating evidence transfer on a realistic use case scenario or domain specific application. Even though evidence transfer proved to be fairly robust during all experiments, estimating label noisy evidence or uncertain evidence might help during the procedure of optimisation and hyperparameter finetuning.

\begin{table*}[h]
\caption{Experimental evaluation of evidence transfer for the CIFAR-10 dataset. $W$ represents the width of the categorical variable that is synonymous to the number of classes of the auxiliary task. $N$ represents the samples size of evidence compare to $M$ primary dataset size. For class incomplete evidence, the left side of ``$\rightarrow$'' represents the original number of classes of the auxiliary task, with the right side representing only the available selected number of auxiliary task classes. Width 3 real evidence is a superset of the labelset that has 3 groups called: ``Vehicles'', ``Pets'' and ``Wild animals'', while width 4 breaks down the ``Wild animals'' into a subset of two groups. Width 5 also breaks down ``Vehicles'' into a subgroup of two groups.}
\label{tab:cifar} \centering %
\subtable[Sample percent incompleteness]{
    \begin{tabular}{|c|c|c|}
    \hline 
    \multirow{1}{*}{Configuration} & ACC (\%) & NMI (\%)\tabularnewline
    \hline 
    Baseline & 22.84 & 13.42\tabularnewline
    \hline 
    Real evidence (w: 3, M=0.1 {*} N) & 25.58 (+2.74) & 16.29 (+2.87)\tabularnewline
    \hline 
    Real evidence (w: 3, M=0.3 {*} N) & 30.81 (+7.97) & 26.47 (+13.06)\tabularnewline
    \hline 
    Real evidence (w: 3, M=N) & 37.34 (+14.50) & 46.24 (+32.82)\tabularnewline
    \hline 
    Real evidence (w: 4, M=0.1 {*} N) & 24.85 (+2.01) & 15.73 (+2.31)\tabularnewline
    \hline 
    Real evidence (w: 4, M=0.3 {*} N) & 32.78 (+9.95) & 27.36 (+13.95)\tabularnewline
    \hline 
    Real evidence (w: 4, M=N) & 43.14 (+20.30) & 54.81 (+41.39)\tabularnewline
    \hline 
    Real evidence (w: 5, M=0.1 {*} N) & 24.95 (+2.12) & 15.76 (+2.35)\tabularnewline
    \hline 
    Real evidence (w: 5, M=0.3 {*} N) & 28.44 (+5.60) & 21.56 (+8.15)\tabularnewline
    \hline 
    Real evidence (w: 5, M=N) & 62.34 (+39.50) & 64.92 (+51.50)\tabularnewline
    \hline 
    Real evidence (w: 10, M=0.1 {*} N) & 23.99 (+1.15) & 14.84 (+1.43)\tabularnewline
    \hline 
    Real evidence (w: 10, M=0.3 {*} N) & 33.94 (+11.11) & 24.03 (+10.61)\tabularnewline
    \hline 
    Real evidence (w: 10, M=N) & 91.97 (+69.13) & 83.06 (+69.64)\tabularnewline
    \hline 
     2 Real evidence (w: 3,4 - M=0.1 {*} N) & 27.08 (+4.25) & 17.87 (+4.45)\tabularnewline
    \hline 
     2 Real evidence (w: 3,4 - M=0.3 {*} N) & 31.99 (+9.16) & 29.18 (+15.77)\tabularnewline
    \hline 
     2 Real evidence (w: 3,4 - M=N) & 52.86 (+30.02) & 61.44 (+48.02)\tabularnewline
    \hline 
     3 Real evidence (w: 3,4,5 - M=0.1 {*} N) & 28.89 (+6.05) & 19.72 (+6.30)\tabularnewline
    \hline 
     3 Real evidence (w: 3,4,5 - M=0.3 {*} N) & 42.95 (+20.11) & 37.90 (+24.48)\tabularnewline
    \hline 
     3 Real evidence (w: 3,4,5 - M=N) & 64.75 (+41.91) & 74.23 (+60.81)\tabularnewline
    \hline 
    \end{tabular}
}
\subtable[Class incompleteness/biasing]{
    \begin{tabular}{|c|c|c|}
    \hline 
    \multirow{1}{*}{Configuration} & ACC (\%) & NMI (\%)\tabularnewline
    \hline 
    Baseline & 22.84 & 13.42\tabularnewline
    \hline 
    Real evidence (w: 3) & 37.34 (+14.50) & 46.24 (+32.82)\tabularnewline
    \hline 
    Real evidence (w: 3 $\rightarrow$ 2) & 29.16 (+6.32) & 28.74 (+15.33)\tabularnewline
    \hline 
    Real evidence (w: 3 $\rightarrow$ 1) & 22.77 (-0.07) & 13.42 (+0.00)\tabularnewline
    \hline 
    Real evidence (w: 4) & 43.14 (+20.30) & 54.81 (+41.39)\tabularnewline
    \hline 
    Real evidence (w: 4 $\rightarrow$ 3) & 33.18 (+10.34) & 38.52 (+25.10)\tabularnewline
    \hline 
    Real evidence (w: 4 $\rightarrow$ 2) & 29.87 (+7.03) & 26.94 (+13.52)\tabularnewline
    \hline 
    Real evidence (w: 5) & 62.34 (+39.50) & 64.92 (+51.50)\tabularnewline
    \hline 
    Real evidence (w: 5 $\rightarrow$ 4) & 40.28 (+17.44) & 52.37 (+38.95)\tabularnewline
    \hline 
    Real evidence (w: 5 $\rightarrow$ 3) & 29.70 (+6.86) & 35.38 (+21.96)\tabularnewline
    \hline 
    Real evidence (w: 10) & 91.97 (+69.13) & 83.06 (+69.64)\tabularnewline
    \hline 
    Real evidence (w: 10 $\rightarrow$ 9) & 87.70 (+64.86) & 83.47 (+70.05)\tabularnewline
    \hline 
    Real evidence (w: 10 $\rightarrow$ 8) & 72.10 (+49.27) & 67.45 (+54.04)\tabularnewline
    \hline 
    2 Real evidence (w: 3,4) & 52.86 (+30.02) & 61.44 (+48.02)\tabularnewline
    \hline 
     2 Real evidence (w: 3,4 $\rightarrow$ 2,3) & 28.78 (+5.95) & 28.37 (+14.95)\tabularnewline
    \hline 
     2 Real evidence (w: 3,4 $\rightarrow$ 1,2) & 24.38 (+1.54) & 13.79 (+0.38)\tabularnewline
    \hline 
    3 Real evidence (w: 3,4,5) & 64.75 (+41.91) & 74.23 (+60.81)\tabularnewline
    \hline 
     3 Real evidence (w: 3,4,5 $\rightarrow$ 2,3,4) & 29.90 (+7.06) & 29.95 (+16.53)\tabularnewline
    \hline 
     3 Real evidence (w: 3,4,5 $\rightarrow$ 1,2,3) & 26.82 (+3.98) & 14.44 (+1.03)\tabularnewline
    \hline 
    \end{tabular}
}
\end{table*}

\section*{Acknowledgments}
This work has been supported by the Industrial Scholarships program of Stavros Niarchos Foundation.

\bibliographystyle{IEEEtran}
\bibliography{IEEEabrv,main}

\end{document}